\documentclass[11pt]{article}

\usepackage{amsfonts,epsfig,graphicx}
\usepackage{amsmath,amssymb,amsthm}
\usepackage{graphics}
\usepackage[authoryear]{natbib}
\usepackage{color}
\usepackage[ruled]{algorithm2e}
\usepackage{epstopdf}
\usepackage{algorithmic}
\usepackage{enumerate}
\usepackage[toc,page]{appendix}
\usepackage{listings}
\usepackage[bookmarks=true,colorlinks,citecolor=blue,urlcolor=blue]{hyperref}

\makeatletter
\long\def\@makecaption#1#2{
        \vskip 0.8ex
        \setbox\@tempboxa\hbox{\small {\bf #1:} #2}
        \parindent 1.5em  %% How can we use the global value of this???
        \dimen0=\hsize
        \advance\dimen0 by -3em
        \ifdim \wd\@tempboxa >\dimen0
                \hbox to \hsize{
                        \parindent 0em
                        \hfil 
                        \parbox{\dimen0}{\def\baselinestretch{0.96}\small
                                {\bf #1.} #2
                                } 
                        \hfil}
        \else \hbox to \hsize{\hfil \box\@tempboxa \hfil}
        \fi
        }
\makeatother

\begin{document} 

\begin{center}

{\LARGE{\bf{{Random Polyhedral Scenes: \\An Image Generator for Active Vision System Experiments}}}}

  \vspace{1cm}

  {\large
\begin{tabular}{ccccc}
Markus D. Solbach && Stephen Voland \cr Jeff Edmonds && John K. Tsotsos
\end{tabular}
}

  \vspace{.5cm}

  \texttt{\{solbach, jeff, tsotsos\}@eecs.yorku.ca} \\

  \vspace{.5cm}

  {\large Department of Electrical Engineering and Computer Science
} \\
\vspace{.1cm}

  {\large York Univeristy, Canada} \\

  \vspace{.5cm}

%%%%%%%%%%%%%%%%%%%%%%%%%%%%%%%%%%%%%%%%%%%%%%%%%%%%%%%%%%%%%%%%%%%%%%%%%%

\begin{abstract}
We present a \textit{Polyhedral Scene Generator} system which creates a random scene based on a few user parameters, renders the scene from random view points and creates a dataset containing the renderings and corresponding annotation files. We hope that this generator will enable research on how a program could parse a scene if it had multiple viewpoints to consider. For ambiguous scenes, typically people move their head or change their position to see the scene from different angles as well as seeing how it changes while they move; this research field is called active perception. 
The random scene generator presented is designed to support research in this field by generating images of scenes with known complexity characteristics and with verifiable properties with respect to the distribution of features across a population. Thus, it is well-suited for research in active perception without the requirement of a live 3D environment and mobile sensing agent, including comparative performance evaluations. The system is publicly available at \url{https://polyhedral.eecs.yorku.ca}.
\end{abstract}
\end{center}

%%%%%%%%%%%%%%%%%%%%%%%%%%%%%%%%%%%%%%%%%%%%%%%%%%%%%%%%%%%%%%%%%%%%%%%%%

\begin{figure}[!ht]
\includegraphics[width=1.0\textwidth]{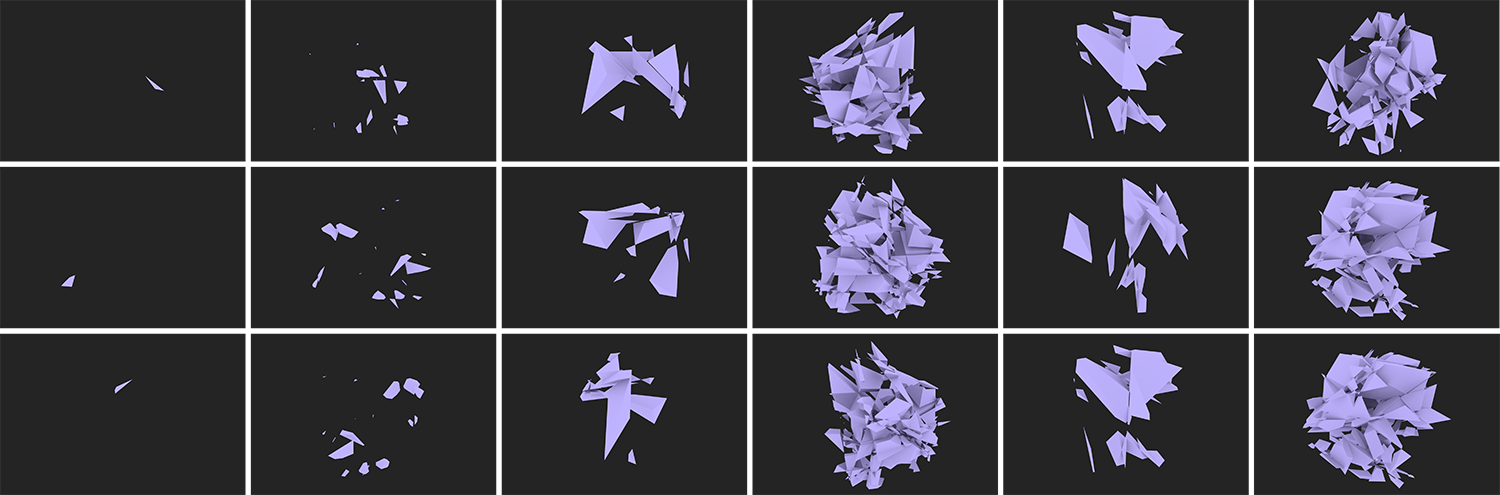}
\centering
\caption{Six different scenes (left to right) rendered from three different views (top to bottom). First row: 1 object. Second row: 18 objects and scene layout set to \textit{separate}. Third row: 18 objects and scene layout set to \textit{touching}. Fourth row: 180 objects and scene layout set to \textit{separate}. Fifth row: 18 objects and scene layout set to \textit{intersecting}. Second row: 100 objects and scene layout set to \textit{intersecting}. More information in Section \ref{sec:UI}.}
\end{figure}

\section{Introduction}\label{sec:Intro}
The system described here provides two main features; (1) it lets a user create a dataset of random views of either a generated polyhedral scene or an uploaded scene and (2) it enables a user to render further views, either with a randomly set camera pose or with a pre-defined camera pose, on-demand.
The development of this random scene generator has a significant context and motivation, which are overviewed now. 
\\\cite{kirousis1985complexity} proved that given a drawing of straight lines depicting a set of opaque polyhedra on the plane: 1) It is NP-Complete, given an image, to tell whether it has a legal labelling; and 2) It is NP-Complete, given an image, to decide whether it is realizable as the projection of a scene. They also proved that the labeling problem has polynomial complexity for line drawings of Legoland (orthohedral) scenes (i.e., scenes made of objects whose 3D edges can only have one of three possible orthogonal directions). This suggests that the brain may exploit geometrical regularities often found in natural scenes. Another possible explanation, which was offered by \cite{kirousis1985complexity}, is that the distribution of natural scenes might be such that the average-case complexity for the set of line drawings extracted from real scenes is polynomial, unlike the complexity for general line drawings. Finally, they conjectured that scenes with shadows or non-trihedral junctions are also very hard (if not much harder).
\\\cite{tsotsos1989complexity} published two theorems addressing the problem of Visual Search, showing that Unbounded Visual Search is NP-Complete (Unbounded denotes that no target-based guidance can be used) while Bounded Visual Match has linear time complexity. This demonstrated one of the points made by \cite{kirousis1985complexity}, namely, that knowledge of the scene reduces the complexity of this particular problem.
\\\cite{parodi1998empirically,parodi1996average} analyzed the complexity of labeling the \cite{kirousis1985complexity} line drawings empirically. The choice of domain takes advantage of the fact that worst-case results are known to be the same as for the Unbounded Visual Search problem; the problem is NP-Complete as described in the previous section. To tackle this issue, they needed to devise a method to generate random instances of polyhedral scenes. Such a method enabled them to generate random line drawings that are guaranteed to be the projection of an actual scene. They then tested several different search methods to label the constructed line drawings. The labeling problem in this case is not the problem of deciding whether a labeling exists, but that of finding a particular labeling. This problem resembles more closely that which the brain solves well, namely the extraction of 3D information from a drawing.  The experiments revealed that the computational complexity of labeling line drawings is, in the median case, exponential in its complexity for the blind, simple depth-first labeling method. On the other hand, it is linear in the number of junctions when using an informed search method, where the information allows knowledge of the task domain to guide processing. The informed search was tested on a much larger set of scenes. Interestingly, there is increasing variability in computational time needed as the number of junctions increases. This is probably due to the highly constrained nature of the labeling problem for trihedral scenes. Although it is possible to construct line drawings containing components that are difficult to label, randomization in the construction of scenes makes these components unlikely to appear. This empirical evidence suggests that both worst-case and median-case analysis lead to the same conclusion: The application of task knowledge guides the processing and converts an exponential complexity problem into a linear one. The shaping or modulation of processing using task-specific knowledge is a major attentive mechanism.
\\The results have important implications. They first tell us that the pure data-directed approach to vision (and in fact to perception in any sensory modality) is computationally intractable in the general case. They also tell us that bounded visual search takes time linearly dependent on the size of the input, something that has been observed in a huge number of experiments (for an overview, see \cite{wolfe1998visual}). Even small amounts of task guidance can turn an exponential problem into one with linear time complexity.
\\These results all use a single image of a 3D scene as input, and thus lead to a question: do additional views change the basic NP-Completeness result of \cite{kirousis1985complexity}? How many views? Which views? How is the next view determined? Such questions naturally fall in the realm of Active Perception (\cite{bajcsy2016revisiting}). This idea has been around for decades, as can be seen from the quote by \cite{popplestone1971relational}: "...consider the object recognition program in its proper perspective, as a part of an integrated cognitive system. One of the simplest ways that such a system might interact with the environment is simply to shift its viewpoint, to walk round an object. In this way, more information may be gathered and ambiguities resolved.  A further, more rewarding operation is to prod the object, thus measuring its range, detecting holes and concavities. Such activities involve planning, inductive generalization, and indeed, most of the capacities required by an intelligent machine."
\\Active Perception by definition takes place in a live 3D environment where the perceiver has the ability to choose what to sense and how to sense it and where those choices depend on the current state of interpretation. Alternatively, it needs a source of images of a scene that cover all possible viewpoints. This seems an impossibility; however, a random scene generator that can generate on demand given imaging geometry parameters can provide exactly this.
\\The random scene generator we present here was designed to support research into active perception by generating images of scenes with known complexity characteristics and with verifiable properties with respect to the distribution of features across a population (as described in \cite{parodi1998empirically}). It is thus well-suited as a foundation for research in active perception without the requirement of a live 3D environment and mobile sensing agent as well as for comparative performance evaluations.
\\As stated at the top of this section, the system provides two main features; (1) it enables a user create a dataset of random views of either a generated polyhedral scene or an uploaded scene and (2) it allows a user to render further views, either with a randomly set camera pose or with a predefined camera pose, on-demand. We now provide more detail on these functionalities.

\section{System Overview}
\begin{figure}[!ht]
\includegraphics[width=1.0\textwidth]{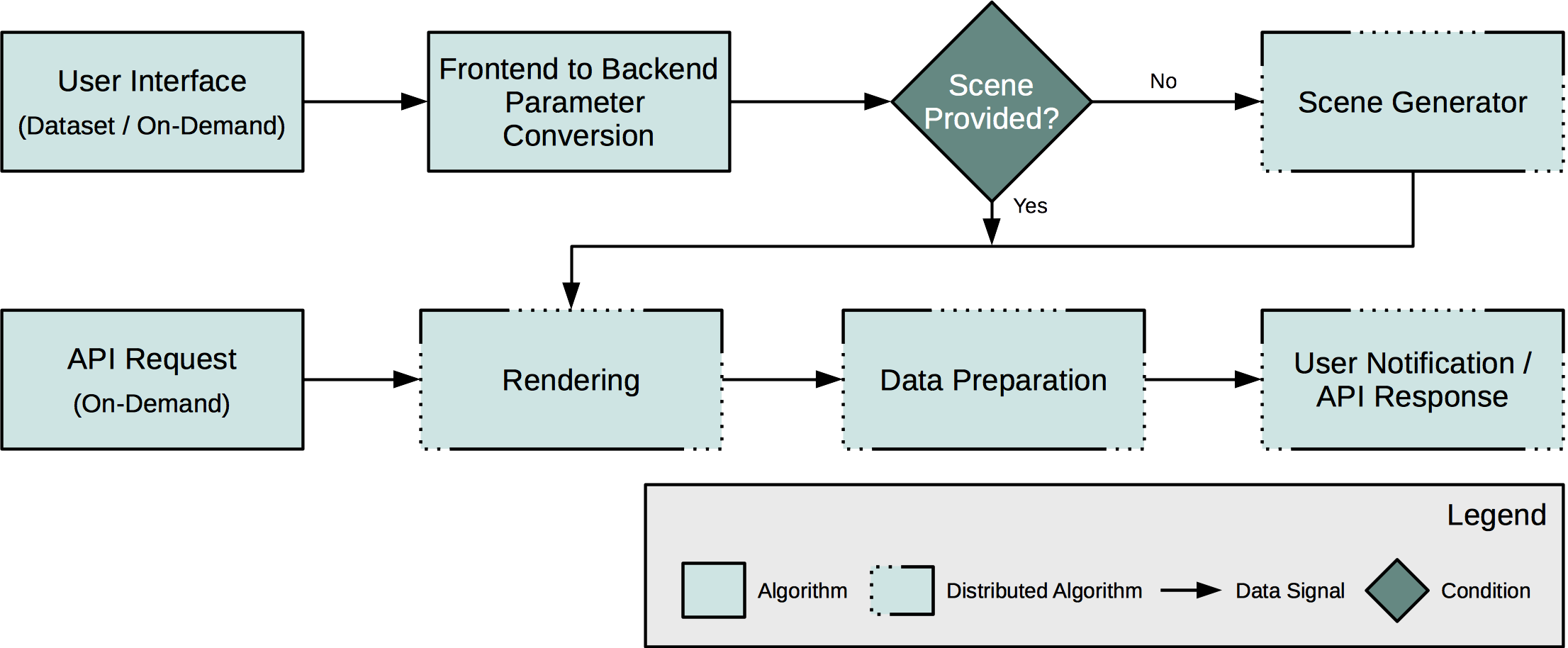}
\centering
\caption{Data flow chart of the entire system. The only parts exposed to the user is the \textit{User Interface} in form of a web page and the \textit{API} --- referred to as the frontend. All remaining parts are considered the backend and are not public.}
\label{fig:sys}
\end{figure}

The structure of the system can be seen in Figure \ref{fig:sys}. The only parts exposed to the user is the \textit{API Request} and the \textit{User Interface} --- referred to as the frontend. The \textit{User Interface} is realized as a web page, serves both features of the system to create a new dataset and process a on-demand request. The \textit{User Interface} is further explained in Section \ref{sec:UI}. The random scene generator also provides an application programming interface (API) which serves the on-demand feature solely. If you are interested in how to use the API, please go to Section \ref{sec:API}.
\\Algorithm \textit{Frontend to Backend Parameter Conversion} is needed to translate between the \textit{User Interface} parameters (\textit{Number of Objects} and \textit{Scene Layout}), to the parameters algorithm \textit{Scene Generator} uses (\textit{Number of Planes} and \textit{Probability of Intersection}). The \textit{Scene Generator} algorithm randomly generates the scene based on the given parameters (further explained in Section \ref{sec:SG}). 
\\The \textit{Rendering} algorithm uses the Python command line interface of Blender to render the scene from different angles (further information in Section \ref{sec:Rendering}). 
\\After the scene is rendered, the data is packed and prepared to be available as a download link and the user is notified via email. Further information about the \textit{Data Preparation} algorithm in Section \ref{sec:DP}.
\\In terms of the overall software architecture, the algorithms \textit{Scene Generator}, \textit{Rendering}, \textit{Data Perparation} and \textit{User Notification} are realised in a distributed fashion using a scalable, non-blocking web server and web application framework. This allows us to distribute the workload conveniently to different machines.

\section{User Interface}\label{sec:UI}
As mentioned in Section \ref{sec:Intro} the \textit{User Interface} serves both features of the system. On one hand, it is the crucial part to create the dataset and obtain an ID. The ID is mandatory to use on the other hand the on-demand feature of the system which lets the user generate further views of the created scene.

\subsection{Dataset Creation}
A screenshot of the \textit{User Interface} to create a dataset can be found at the Appendix in Figure \ref{fig:UI_data}.
\\In this configuration, the user has the choice to either use the polyhedral scene generator, further explained in Section \ref{sec:SG}, or upload a scene. The uploaded scene has to be saved as a Wavefront (.obj) file. In case the user wants to create a plyhedral scene, the system accepts four parameters; \textit{Number of Objects}, \textit{Scene Layout}, \textit{Lighting Conditions} and \textit{Number of Views}:

\begin{itemize}
  \item \textbf{Number of Objects} --- defines the number of objects generated and placed in the scene.
  \item \textbf{Scene Layout} --- describes how the objects will be placed within the scene; distinct from each other, allowed to touch each other or allowed to intersect.
  \item \textbf{Lighting Conditions} --- describes the two different lighting conditions. Homogeneous lighting places six light sources in the scene pointing from each side at the scene. In this setup, less shadows are visible. Fixed spotlight places one spotlight in the scene. In this setup, richer shadows are visible.
  \item \textbf{Number of Views} --- defines the number of renderings performed from random positions selected on a sphere around the scene.
  \item \textbf{Email address} --- the email address to which the download link will be provided to.
\end{itemize}

The user interface provides the user with necessary information to select appropriate parameters, gives example renderings of different parameter settings and also explains the rendering setup. In detail, the rendering setup is explained in Section \ref{sec:Rendering}.
\\As an example, let us assume that a user wants to create a scene with two separate polyhedral objects with a single fixed light source and with 100 random views of this scene. In this case, a user would choose the following parameters:
\begin{itemize}
  \item \textbf{Number of Objects} --- 2
  \item \textbf{Scene Layout} --- Separate
  \item \textbf{Lighting Conditions} --- Fixed spotlight
  \item \textbf{Number of Views} --- 100
\end{itemize}

\subsection{On-Demand Request}
A screenshot of the \textit{User Interface} to execute an on-demand request can be found at the Appendix in Figure \ref{fig:UI_demand}. Important to note at this point is that this feature is only available if the user already created a dataset, hence received an ID.
\\In this configuration the user able to render further scenes of the previously set up scene. The user can choose to either render another random view or to place the camera. To enter the camera pose easier, the web page provides a three-dimensional toy scene in which the user can set the camera (green ball) to a certain position by using the mouse. In this case, the camera will be always pointed at the center of the scene. Whenever the user picks a new position of the camera, the camera parameter values will update accordingly. Another option is to set the camera parameters by hand. This also allows to point the camera at an abritrary position in the scene.
\\In this configuration, the user has to set 4 parameters; \textit{ID}, \textit{Random or defined Camera}, \textit{Camera Pose} and \textit{Lighting Conditions}:

\begin{itemize}
  \item \textbf{ID} --- defines the ID of a previously generated scene. The ID is provided in the EMail of the Dataset Creation. In case the user does not have an ID yet, the user is asked to submit a \textit{Data Creation} request first. 
  \item \textbf{Random or defined Camera} --- describes how the camera will be placed. \textit{Random} will place the camera randomly in the scene pointing at the center of the scene. When set to \textit{Defined}, six additional input fields are displayed to define the camera pose.
  \item \textbf{Camera Pose} --- defines the camera pose as $(x, y, z)$ translation and $(qw, qx, qy, qz)$ orientation in quaternion form. The user has total flexibility of the camera pose. The web page will provide a three-dimensional toy scene in which the user can set the camera (green ball) to a certain position by using the mouse. If the user decides to pick the camera pose with the toy scene, the camera will be always pointed at the center of the scene.
  \item \textbf{Lighting Conditions} --- describes the two different lighting conditions. Homogeneous lighting places six light sources in the scene pointing from each side at the scene. In this setup, less shadows are visible. Fixed spotlight places one spotlight in the scene. In this setup, richer shadows are visible.
\end{itemize}
As an example, let us assume that a user wants to generate further views from a previously generated scene, choose homogeneous lighting, set the camera to a certain position and use the web page to do so. In this case, a user would choose the following parameters:
\begin{itemize}
  \item \textbf{ID} --- ID of scene
  \item \textbf{Random or defined Camera} --- Defined
  \item \textbf{Camera Pose} --- Pick the camera location in the \textit{Camera Setter} section of the web page. This should fill-in the information for $(x, y, z)$ and $(qw, qx, qy, qz)$.
  \item \textbf{Lighting Conditions} --- Homogeneous lighting 
\end{itemize}

\section{Scene Generator}\label{sec:SG}
The \textit{Scene Generator} is the part of the system that generates the scene based on the given parameters and saves the scene as a Wavefront .obj file. 
\\The \textit{Scene Generator} starts by generating the requested number of planes in vector format (point and normal vector). To produce the point it randomly generates $x$, $y$, and $z$ coordinates in the interval $[-1, 1]$, and then discards the point and tries again if the resulting point is not within a sphere of radius 1 centered on the origin. The normal vector is a random vector from the origin to the unit sphere, composed of two angles. The first, $\phi$, is simply a unit vector in the $xy$-plane; its angle is randomly chosen from [0, 2$\pi$]. The second, $\theta$, represents the elevation in the $z$-plane.  Its interval is [-$\pi$, $\pi$], but it can't simply be chosen at random from this interval because a random vector is much more likely to point to the equator than to a pole.  Instead, $\theta$ is generated by taking arcos of a random number in the interval $[-1, 1]$.
\\From this start, scene generation is straightforward. A line is created everywhere that two planes intersect within the unit sphere, and a point where three planes intersect. The line segments delineated by the points are checked to see whether they make up a polygon, and the polygons in turn are checked to see whether they make up a polyhedron. Finally, each polyhedron is randomly determined to be visible or not based on the user-selected \textit{Probability of intersection} parameter.
\\The algorithm is written in Java and implements the proposed algorithm by \cite{parodi1998empirically}. For further information please refer to \cite{parodi1998empirically}.
\\An extension was added to \cite{parodi1998empirically} to assure that the generated scene consists of the right number of objects. Since the scene generation is random, we perform algorithm \cite{parodi1998empirically} multiple times until the scene has the correct number of objects. Module \textit{Frontend to Backend Parameter Conversion} gives good parameters that usually results in the right number of objects after a few runs.

\section{Rendering}\label{sec:Rendering}
Once the scene is generated the rendering is done using Blender\footnote{\url{https://www.blender.org/}}. Blender provides a state of the art rendering engine and a command line interface to automatically render a scene and manipulate the camera using a Python script. Blender offers different types of rendering engines, the most advanced one and the engine we use to render the scene is \textit{Cycles}. \textit{Cycles} is Blender's ray-trace based production render engine and provides realistic material and shading effects.
\begin{figure}[!ht]
\includegraphics[width=1.0\textwidth]{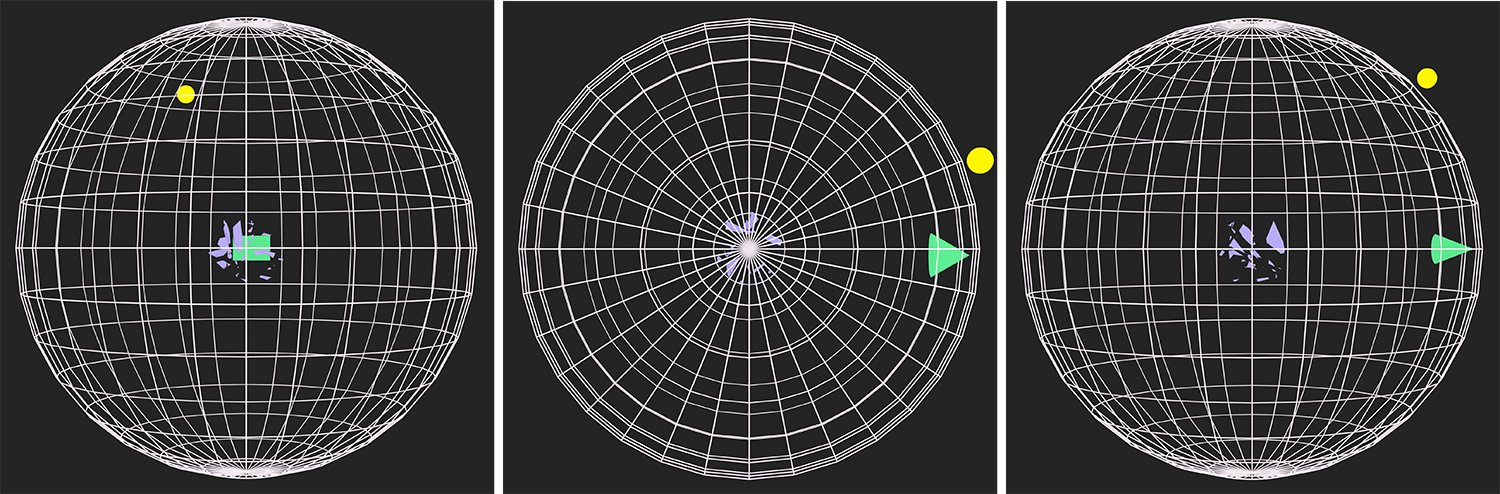}
\centering
\caption{\textbf{Fixed Spotlight:} Illustration of the rendering setup. Yellow illustrates the light source (Point Lamp). Green shows the camera. Purple the random polyhedral scene. White is the sphere on which the camera is randomly positioned to render the purple scene from.}
\label{fig:rendering}
\end{figure}
The Material settings for the scene are set to one ambient material, with an ambient occlusion surface and its color set to $\#BEB3FF$ and alpha to 1.0. The user has the option to choose between two lighting scenarios; one light source (fixed point lamp) with a size of 1.103 and color set to $\#FFFFFF$ or six light sources (hemisphere lamps) with max bounces set to 1024 and color set to $\#FFFFFF$. The first option will result in richer shadows and the second option in a more homogeneously illuminated scene.
\begin{figure}[!ht]
\includegraphics[width=1.0\textwidth]{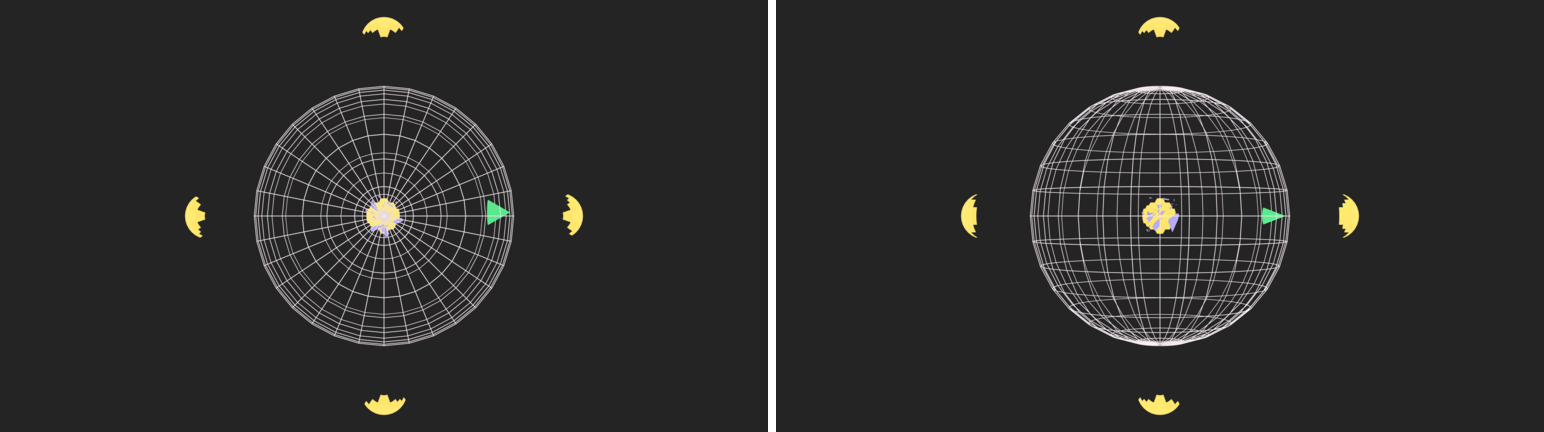}
\centering
\caption{\textbf{Homogeneous Lighting:} Illustration of the rendering setup. Similar to Figure \ref{fig:rendering}, except for yellow illustrates six light sources (Hemisphere Lamp).}
\label{fig:rendering_homo}
\end{figure}
The rendering viewport is a standard camera object of Blender with a perspective lens, focal length of $35mm$, a sensor size of $32mm$ and a F-Stop of $128$. The output resolution is set to $1920px \times 1080px$.
\\The scene's origin is at $(0,\ 0,\ 0)$ and will not exceed a space beyond $[-3,\ 3]$ at each axis. Taking the maximum scene dimension into account, we positioned a virtual sphere with size $[5,\ 5]$ around the scene's origin to place the camera onto (facing point $(0,\ 0,\ 0)$) to render the scene. An illustration of the rendering setup can be seen in Figure \ref{fig:rendering}.
\\To choose random rendering view points on the screen, we followed the algorithm by \cite{marsaglia1972choosing}. This algorithm guarantees to have a uniform distribution over a unit sphere. Other approaches using spherical coordinates $\theta$ and $\phi$ from uniform distribution $\theta \in [0, 2\pi] $ and $\phi \in [0, \pi] $ have the disadvantage that points are picked more likely near the poles \cite{marsaglia1972choosing}.

\section{Data Preparation}\label{sec:DP} 
\begin{figure}[!ht]
\includegraphics[width=1.0\textwidth]{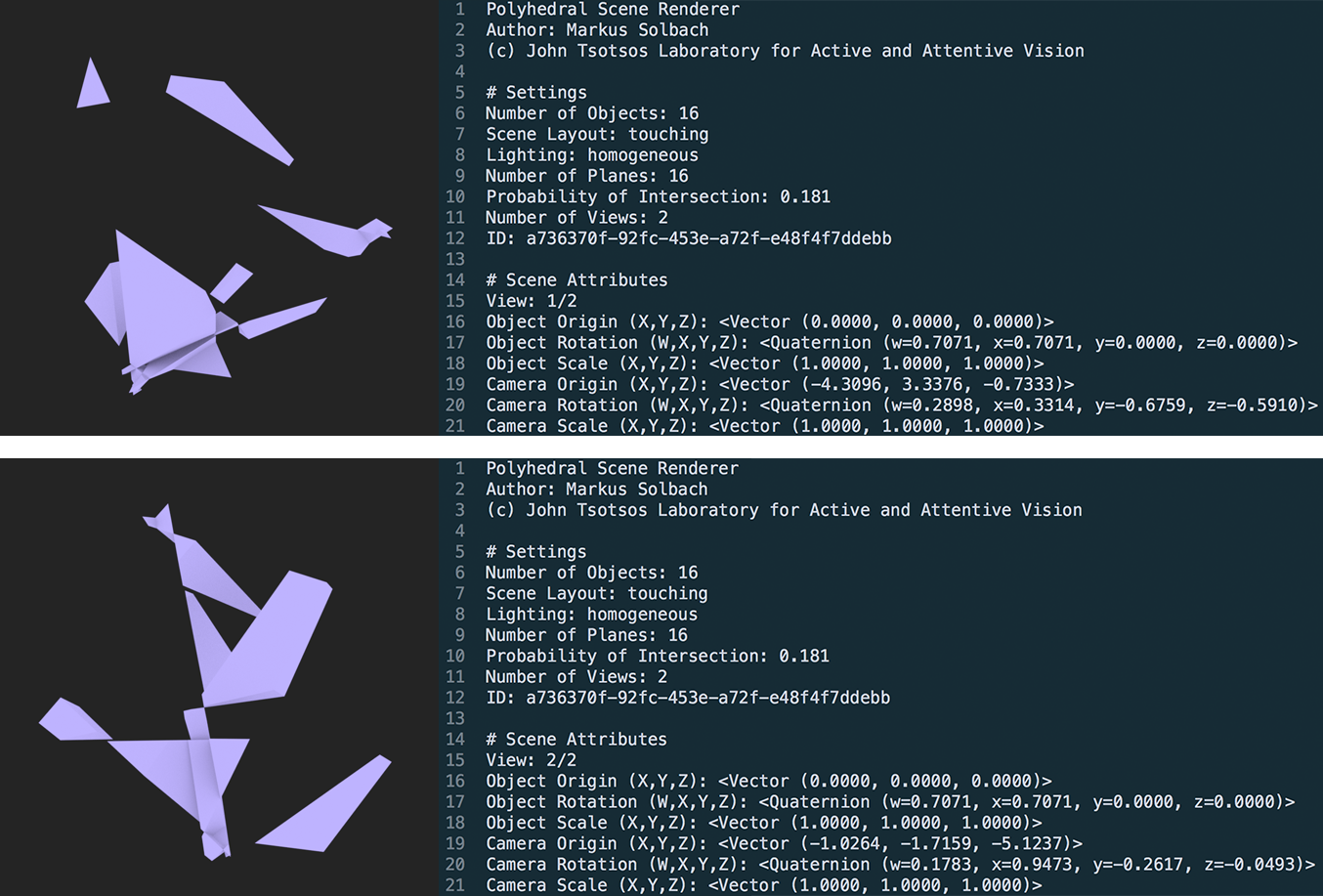}
\centering
\caption{Example data set created by the \textit{Polyhedral Scene Generator}. Each row shows one image and annotation file pair.}
\label{fig:data}
\end{figure}

The algorithm \textit{Data Preparation} creates a downloadable data folder with the renderings and also the annotation files for each rendering. An example can be seen in Figure \ref{fig:data}. The Figure shows the rendering on the left and the corresponding annotation file on the right. The annotation file consists --- in addition to some meta information --- of the chosen user settings, including the calculated parameters with which the scene generator was run and the scene attributes. The scene attributes depict the current number of the view (line $15$), object transformation (line $16-18$), where object, in this case, stands for the entire polyhedral scene. The camera transformation is described in line $19-21$. Corresponding annotation and image file share the same file name and should be easy to identify. Line $12$ shows the ID which is necessary to use the API (see Section \ref{sec:API}).

\section{Application Programming Interface}\label{sec:API}
Besides a graphical user interface in the form of a webpage, the polyhedral scene generator offers an application programming interface (API). To use the API, the user needs to create a polyhedral scene through the web interface first to specify the scene and obtain an ID. Once achieved the ID, the user has everything to use the API. The API is designed to be as lightweight as possible and basically provides the opportunity to render the scene from either another random or defined viewing angle. To do so, in total three parameters (random viewing angle) or four parameters (defined viewing angle) have to be set. The parameters are as follows:
\begin{itemize}
  \item \textbf{id} (\textit{string}) --- ID of the polyhedral scene as provided in the annotation file (line 12).
  \item \textbf{lighting} (\textit{string}) --- \textit{fixed} or \textit{homogenous}. \textit{Fixed}: as described earlier with one point lamp positioned at $(-4.6, 1.763, -4.493)$. \textit{Homogenous}: six hemisphere lamps pointing from each side to the object to provide a more homogeneous lighting.
  \item \textbf{random} (\textit{boolean}) --- This parameter defines whether the camera is going to be set at a certain position (\textit{false}) or positioned randomly in the scene (\textit{true}).
  \item \textbf{camera} (\textit{float}) --- Ignored if \textit{random} is set to \textit{true}. Holds a JSON object with seven elements describing the six dimensional pose of the camera with $[x, y, z]$ for translation and $[qw, qx, qy, qz]$ to describe the rotation using quaternion.
\end{itemize}

The response of the API call will be a JSON in the structure as seen in Listing \ref{ls:resp}.

\begin{lstlisting}[language=Python, caption={Example response}, label={ls:resp}]
{
   "status": "200",
   "image": "goAAAANSUhEUgAAAZAAAADSCAMAAABT..."
}
\end{lstlisting}

Where \textit{status} provides feedback to the user using http status codes and \textit{image} contains the image decoded as a base64 string.
\\An example on how to use the API can be seen in Listing \ref{ls:py} or at \\\url{https://solbach.github.io/polyhedral/}. 

\section{Suggested Usage}
The presented work can be used for different scenarios which need multiple views from a certain scene. Our system provides two main features that enables research in a wide variety of fields in active vision, active perception and attentive vision. The first feature is the ability to create an entire dataset of random views of a known or unknown scene. The second feature is the ability to provide control over the camera to collect further views of the scene. Additional views can be either collected randomly or by choosing the exact camera pose.

\subsection{Dataset Creation --- Scene Setup}
\begin{figure}[!ht]
\includegraphics[width=0.7\textwidth]{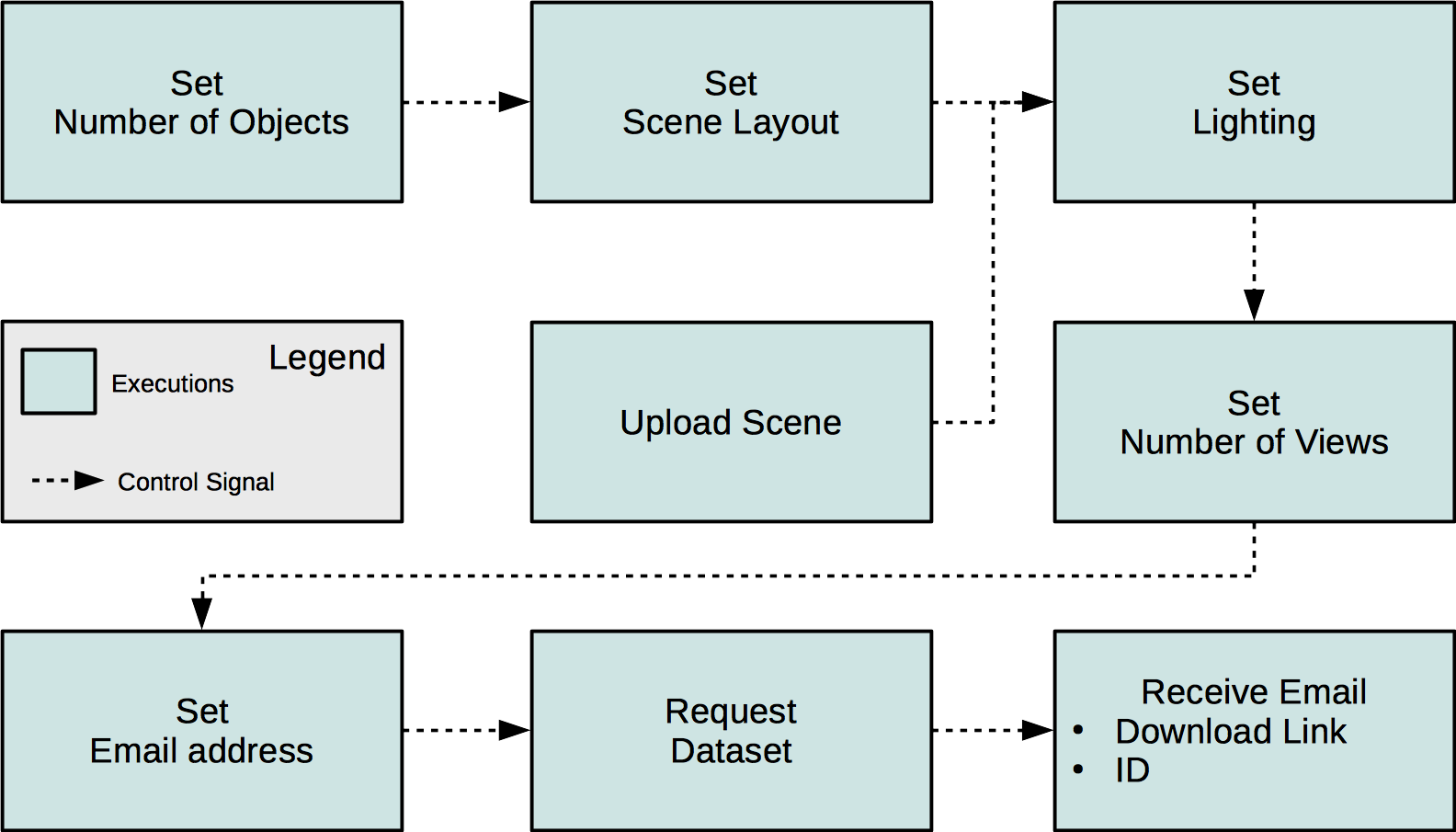}
\centering
\caption{Example of how to use the system for dataset creation. Web page only}
\label{fig:dataset_use}
\end{figure}

The first feature of the system is to provide the capability to create a dataset. Figure \ref{fig:dataset_use} illustrates a classic workflow using the system for this purpose.
The functionality of the dataset creation is only available through the web page and cannot be used through the API. Two ways exist to perform this step; the first way creates a random polyhedral scene by first selecting the number of objects and setting the scene layout, the second possible way is to upload your own scene (Wavefront format) and skips the two settings of way one. Afterwards, the user, regardless using way one or two, sets the lighting condition, number of views, email address and requests the dataset. After a short while the user will receive an email with a download link and ID. The ID is crucial to use the on-demand functionality described next.

\subsection{On Demand Request --- Further Views}
\begin{figure}[!ht]
\includegraphics[width=0.7\textwidth]{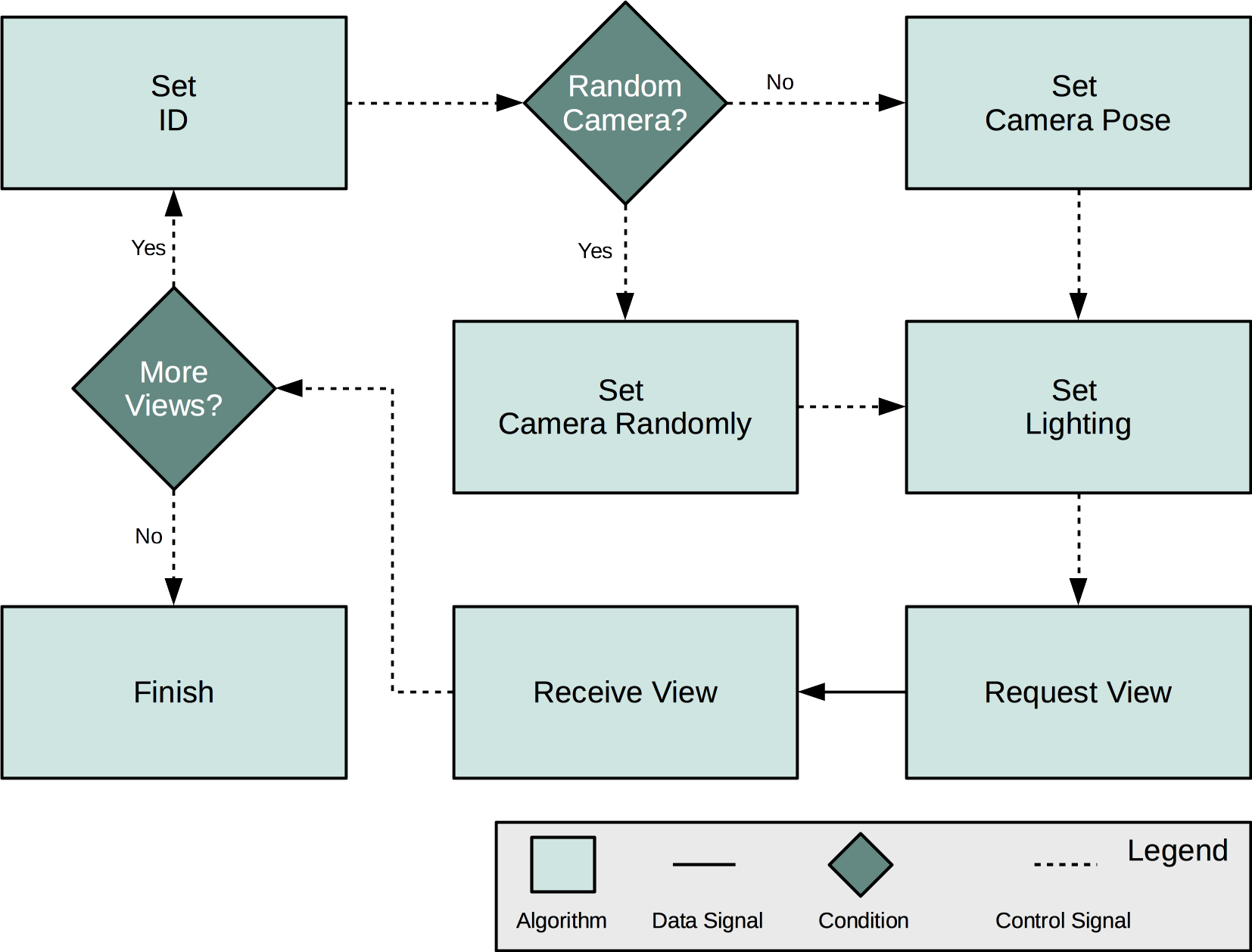}
\centering
\caption{Example of how to use the system for on-demand use. Web page and API}
\label{fig:demand_use}
\end{figure}

The second feature of the system is to provide the capability to create additional views of the previously defined scene. This feature is called \textit{On-Demand Request} and is illustrated in Figure \ref{fig:demand_use}. Besides a web page implementation this feature is also available as a web API.
The user sets first the ID and can choose to use a randomly placed camera or to set the camera by hand. In the first case, the camera will be placed at a random position on a sphere spanning around the scene, pointing at the center of the scene. In the second case, the user sets the camera pose. Random camera or set camera, in both cases the user is able to select the lighting conditions. The time between requesting and receiving the additional view depends on the workload of the system and can vary from 10s - 30s. Chose the user a randomly placed camera, the system will provide the used camera pose together with the image. 
\\After receiving the view the user can decide to obtain further views (going back to \textit{set ID}) or to finish.

\section{Conclusion}
We present a system that is well-suited for research in active perception without the requirement of a live 3D environment and mobile sensing agent as well as for comparative performance evaluations. The funcionality of this system is twofold. Firstly, it provides the capability to create a random polyhedral scene with known complexity characteristics and with verifiable properties with respect to the distribution of features across a population. Secondly, it provides the functionality to render further views, either with a randomly set camera pose or with a pre-defined camera pose, on-demand. Additionaly, the system can be also used with existing scenes which can be uploaded through the web page.
\\Furthermore, we want to point out the following:
\begin{itemize}
  \item We provide no guarantees for system performance.
  \item The system is provided as is.
  \item Use of this system is permitted for research purposes only and commercial use is not permitted.
  \item For comments, problems and questions, please send an email to 
  \\\url{polyhedral@eecs.yorku.ca} (replies are not guaranteed).
  \item If You use any aspect of this system, we respectfully request that all relevant publications that result from any use of this paper or system cite this paper.
\end{itemize}

\bibliographystyle{apalike} 
\bibliography{bib.bib}

\newpage

\begin{appendix}

\section{Appendix}

\begin{figure}[!ht]
\includegraphics[height=0.88\textheight]{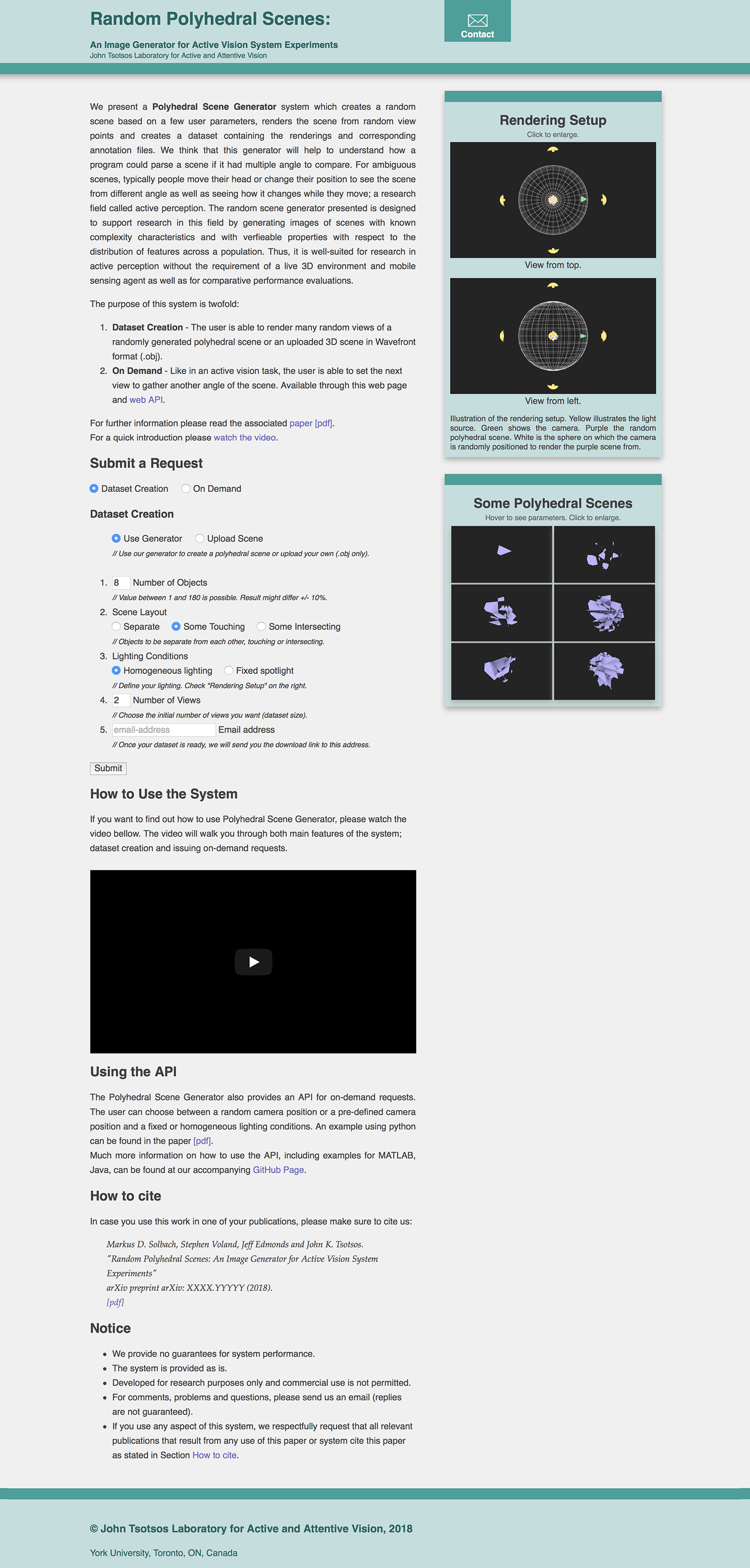}
\centering
\caption{\textbf{Dataset Request:} User Interface of the \textit{Polyhedral Scene Generator}. }
\label{fig:UI_data}
\end{figure}

\begin{figure}[!ht]
\includegraphics[height=0.88\textheight]{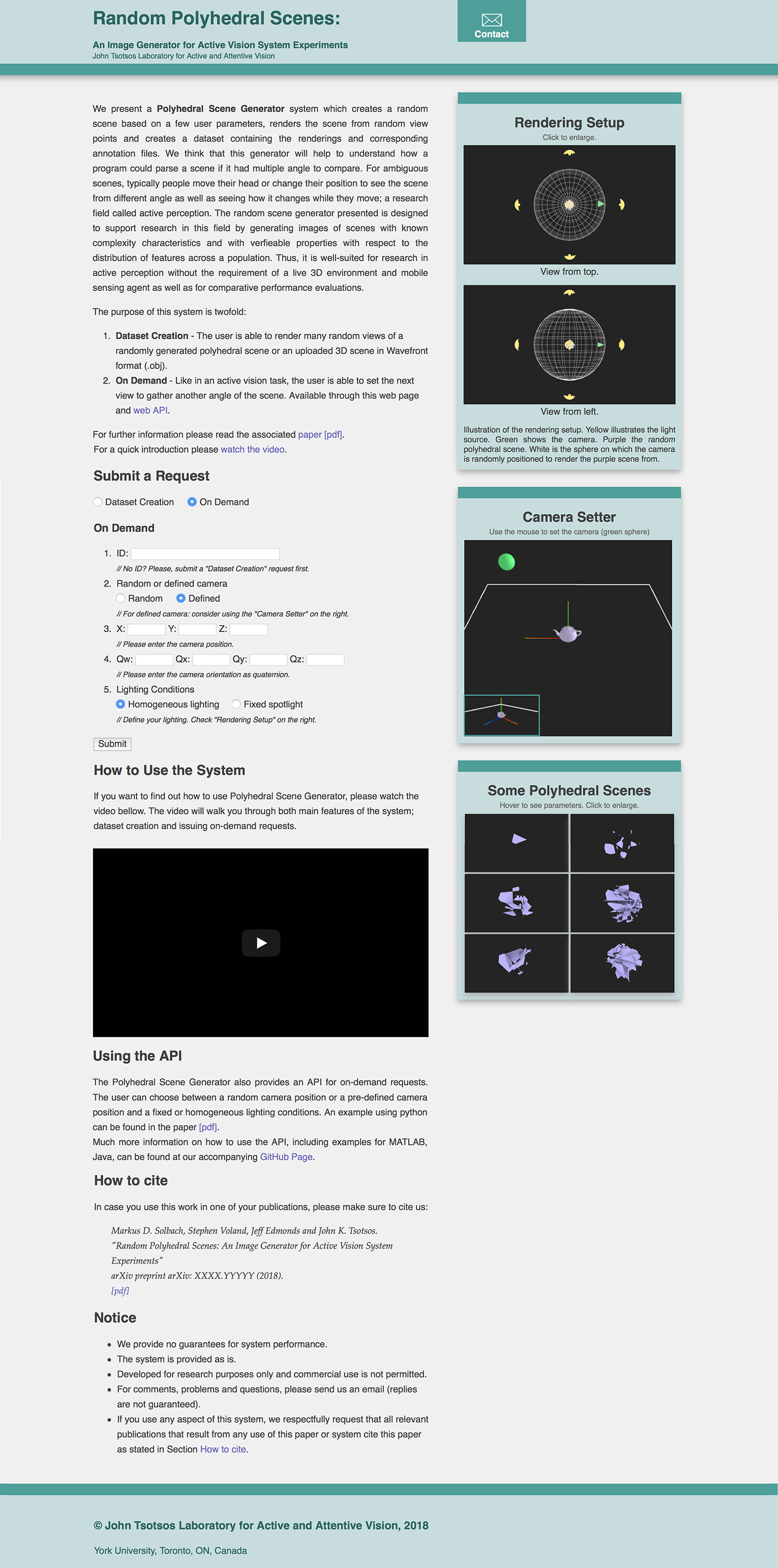}
\centering
\caption{\textbf{On Demand Request:} User Interface of the \textit{Polyhedral Scene Generator}. }
\label{fig:UI_demand}
\end{figure}

\newpage

\begin{lstlisting}[language=Python, caption={Example Python code for using the API.}, label={ls:py}]
# Author: Markus Solbach (polyhedral@eecs.yorku.ca)
from websocket import create_connection
import io, sys, json, base64
from json import dumps
try:
    from PIL import Image
except ImportError:
    print("PIL not installed on system. Running \
           lightweight example.")
# Create Connection
ws = create_connection(
          "wss://polyhedral.eecs.yorku.ca/api/")
# Prepare Data
parameter = {
    'ID':'YOUR ID HERE',
    'light_fixed':'true',
    'random_cam': 'true',
    'cam_x':-0.911,
    'cam_y':1.238,
    'cam_z':-4.1961,
    'cam_qw':-0.0544,
    'cam_qx':-0.307,
    'cam_qy':0.9355,
    'cam_qz':0.16599
}
json_params = dumps(parameter, indent=2)
ws.send(json_params)
while True:
    result = json.loads(ws.recv())
    print("Job Status: {0}".format(result['status']))
    if result['status'] == "SUCCESS":
        break
    elif "FAILURE" in result['status'] \
      or "INVALID" in result['status']:
        sys.exit()
image_base64 = result['image']
image_decoded = base64.b64decode(image_base64)
random_cam_param = result['cam_pose']
print(random_cam_param)
fh = open("imageToSave.png", "wb")
fh.write(image_decoded)
fh.close()
if 'PIL' in sys.modules:
    im = Image.open(io.BytesIO(image_decoded))
    im.show()
ws.close()
\end{lstlisting}

\end{appendix}

%%%%%%%%%%%%%%%%%%%%%%%%%%%%%%%%%%%%%%%%%%%%%%%%

\end{document}